\documentclass{esannV2}
\usepackage[dvips]{graphicx}
\usepackage{amssymb,array}
\usepackage{amsmath,tabstackengine}
\usepackage{epstopdf}
\usepackage{url}
\usepackage[font=footnotesize]{subfig}
\usepackage[utf8]{inputenc}
\usepackage[T1]{fontenc}
\usepackage[svgnames]{xcolor}
\usepackage{algorithm2e}
\usepackage{multirow} % added 
\usepackage[numbers]{natbib}
\stackMath

\newlength{\bibitemsep}
\newlength{\bibparskip}
\let\oldthebibliography\thebibliography
\renewcommand\thebibliography[1]{%
  \oldthebibliography{#1}%
  \setlength{\parskip}{\bibitemsep}%
  \setlength{\itemsep}{\bibparskip}%
}

\def\dbR{{\mathrm{I\hskip-2.2pt R}}}

\def\v{{\mathbf{v}}}

\def\x{{\mathbf{x}}}
\def\y{{\mathbf{y}}}

\def\1o{{\mathrm{\bigcirc\hskip-7pt \text{\footnotesize{1}}}}}
\def\2o{{\mathrm{\bigcirc\hskip-7pt \text{\footnotesize{2}}}}}
\def\3o{{\mathrm{\bigcirc\hskip-7pt \text{\footnotesize{3}}}}}
\def\4o{{\mathrm{\bigcirc\hskip-7pt \text{\footnotesize{4}}}}}
\def\5o{{\mathrm{\bigcirc\hskip-7pt \text{\footnotesize{5}}}}}
\def\6o{{\mathrm{\bigcirc\hskip-7pt \text{\footnotesize{6}}}}}
\def\7o{{\mathrm{\bigcirc\hskip-7pt \text{\footnotesize{7}}}}}
\def\8o{{\mathrm{\bigcirc\hskip-7pt \text{\footnotesize{8}}}}}
\def\9o{{\mathrm{\bigcirc\hskip-7pt \text{\footnotesize{9}}}}}

\begin{document}
%style file for ESANN manuscripts
\title{ $L_1$-norm double backpropagation adversarial defense}
%\title{Adversarial defense via double backpropapagation }

%***********************************************************************
% AUTHORS INFORMATION AREA
%***********************************************************************
\author{Ismaila Seck$^{1,2}$, Gaëlle Loosli$^{2,3}$ and St\'ephane Canu$^1$
%
% Optional short acknowledgment: remove next line if non-needed
%\thanks{This work was performed within the framework of .}
%
% DO NOT MODIFY THE FOLLOWING '\vspace' ARGUMENT
\vspace{.3cm}\\
%
% Addresses and institutions (remove "1- " in case of a single institution)
1- Normandie Univ, INSA Rouen, UNIROUEN, UNIHAVRE, LITIS, France\\
\vspace{.01cm}\\
2- UCA - LIMOS UMR 6158 CNRS  \\
Clermont-Ferrand, France\\
\vspace{.01cm}\\
3- PobRun, Brioude, France
}
%***********************************************************************
% END OF AUTHORS INFORMATION AREA
%***********************************************************************

\maketitle

\begin{abstract}
Adversarial examples are a challenging open problem for deep neural networks. We propose in this paper to add a penalization term  that forces the decision function to be flat in some regions of the input space, such that it becomes, at least locally, less sensitive to attacks.
Our proposition is theoretically motivated and shows on a first set of carefully conducted experiments that it behaves as expected when used alone, and seems promising when coupled with adversarial training. 
\end{abstract}

%---------------------------------------------------------------------------------------------
\section{Introduction}
\vspace{-0.7em}
%---------------------------------------------------------------------------------------------
Deep learning algorithms have set the state-of-the-art in several domains among which image classification. 
However,  \cite{Goodfellow2014} showed that it is possible, and relatively easy, to fool Deep Neural Networks by adding to the inputs a particular perturbation. This added perturbation may be such that the disturbed inputs and the original ones are very close according to some metrics, but are assigned to different classes.
%is such that the perturbed inputs and original ones can be very close according to some metric but are classified in different classes. 
Several defense mechanisms have been introduced to prevent such a behavior, but 
their efficiency is limited to specific cases so that the general problem of 
preventing the existence of adversarial examples remains open. 

One idea is to have a constant output over a region around known training points. For a differentiable function, that means an arbitrarily chosen norm of the output's gradient with respect to the input should be 0 or at least as small as possible over that region. 
Our claim is that by using adversarial training and by penalizing the gradient's norm of the output with respect to the input, the robustness of the model can be improved. 
The $L_1$-norm is chosen and this choice will be theoretically motivated by calculus.

\vspace{-0.7em}
%---------------------------------------------------------------------------------------------
\section{Related work}
%---------------------------------------------------------------------------------------------
\vspace{-0.7em}

The adversarial examples were first presented in \citep{szegedy2013intriguing}, and the principle of adversarial training was introduced at the same time. Adversarial training consists in augmenting the dataset with potentially adversarial points. % {\color{red} il manque des mots??}
But it was impractical, since the method used to generate adversarial samples, L-BFGS, was too slow. Adversarial training became more convenient to use with the introduction of Fast Gradient Sign Method (FGSM) \cite{Goodfellow2014}, which is much faster and generates adversarial examples with a good success rate. Moreover, using a first order expansion of Taylor series, adversarial training can be seen as an $\ell_1$-norm penalty of the derivative of the loss with respect to the inputs \citep{simon2018adversarial}. Note that  {\it regularization functional which penalize derivatives of the resulting classifier function are not typically used in deep learning} \citep{hein2017formal}. %Two exceptions are  double backpropagation \citep{drucker1992improving,hochreiter1995simplifying} who implement the idea of finding  large connected regions of the error function called flat minima. 

The idea of penalizing the gradient of the output with respect to the input  was introduced by \cite{drucker1992improving} under the name \textit{double backpropagation} and later used in \cite{hochreiter1995simplifying} to find large connected regions of the error function called flat minima. The ultimate goal was the improvement of the generalization of their models, which differs slightly from our goal here. 
In \textit{double backpropagation}, the $\ell_2$-norm of the loss is penalized. Using the Energy loss function, it was stressed that the penalization of the loss would have little to no effect when the classification is good. To balance out that effect, the multiplicative parameter of the penalization ought to be large enough.%{\color{red} "to correct?": pas sure de comprendre }

The difference between double backpropagation and our gradient penalization is that we penalize the $\ell_1$-norm of the gradient of each output with respect to the input while, in backpropagation, the $\ell_2$-norm of the loss is penalized. Hence, we should not have the problem that occurs when the penalization term is multiplied by a small error vector. Nevertheless, this might not be a problem when using a different loss function. Although empirical evidence show double backpropagation's efficiency to enhance generalization, it is insufficient to defend against adversarial examples. That is what \cite{papernot2016distillation} highlights saying  {\it{limiting sensitivity to infinitesimal perturbation \citep[e.g., using double backpropagation][]{drucker1992improving} only provides constraints very near training examples, so it does not solve the adversarial perturbation problem}}. But there are evidence that coupling that gradient penalty with adversarial training increases the robustness.

%---------------------------------------------------------------------------------------------
\section{Defensive gradient penalty}
%---------------------------------------------------------------------------------------------
\vspace{-0.7em}
 
It has been shown that that maliciously crafted examples can fool the deep learning classifiers and lead them to misclassify  those examples with high confidence. In addition to the adversarial training, a gradient penalization is proposed here to make the models more robust. Let $\x$ be an input image stored in a vector belonging to the input space $\mathcal{X}=[0,1]^d$, where $d$ is the dimension of $\x$ and let $y$ be the target, a one-hot label, associated with $\x$.
 Let $f$, a differentiable function, $f: \mathcal{X} \to \dbR^{c}$ where c is the number of classes, such that the decision function $D(x) = argmax(f(x))$ represents our classifier.
And let $\mathcal{L}(f(\x),\y)$ be a common differentiable loss function.   
 
 For a given input $\x$, a perturbation direction $\v$ and a positive scalar $\varepsilon$,   the first order expansion of the transfer function i-th component, $f_i$, of the neural network is: 
\begin{equation}
    f_i(\x+\varepsilon \v) = f_i(\x) + \varepsilon \v^{\mbox{\tiny T}} \nabla_\x f_i(\x) +
    \circ (\varepsilon\|v\|).
\end{equation}
\noindent Considering the FGSM attack \citep{Goodfellow2014} with  $\v=\mbox{sign}(\nabla_\x \mathcal{L}(f(\x),\y)$, we have $\v^{\mbox{\tiny T}} = \{\pm1\}^d$,  
$|f_i(\x+\varepsilon \v) - f_i(\x)|\approx|\varepsilon \v^{\mbox{\tiny T}} \nabla_\x f_i(\x)| \le \varepsilon \|\nabla_\x f_i(\x)\|_1
= \varepsilon w^{\mbox{\tiny T}}\nabla_\x f_i(\x)$ for $w=$sign$(\nabla_\x f(\x))$.
So that,  in this case, finding weights that minimize sensitivity to infinitesimal perturbation of the input can be done by minimizing the $\ell_1$-norm of the gradient of each component of the transfer function with respect to the input. Our approach is therefore to penalize the $\ell_1$-norm of the gradient of each coordinate $f_i(x)$ not only on the original point but also on the potentially adversarial point generated. If we manage to have 0 as the $\ell_1$-norm of those gradients at two points that are very close, then the output of the classifier is almost constant along the segment joining those two points. Indeed we have: 
\vspace{-0.7em}
\begin{equation}
    0 \leq |f_i(\x+\varepsilon \v) - f_i(\x)| \leq \varepsilon \sup_{t \in [0,\varepsilon]} \|\nabla_\x f_i(\x+t\v)\|_1  
\end{equation} 
\noindent %This equation highlights again that adversarial training can be seen as an $\ell_1$-norm regularization with $\varepsilon$ as an associated coefficient. 
The regularization coefficient being small ($\varepsilon$ has to be small for $\x+\varepsilon \v$ to be considered as an adversarial example) the regularization is not used at its full potential. In order to increase the regularization coefficient, an explicit penalization of the $\ell_1$-norm is added to the loss function. Hence the analogy with \citep{drucker1992improving}, which penalize the $\ell_2$-norm of the gradient of the loss while in this paper the sum of the $\ell_1$-norm of the outputs ($f_1(\x),f_2(\x), \dots, f_c(\x)$) with respect to the input is directly penalized. But, %\textit{ using double backpropagation\citep{drucker1992improving} only provides constraints very near training examples, so it does not solve the adversarial perturbation problem.} \citep{papernot2016distillation},
it is very hard to make derivatives very small using only the penalization of the gradient when penalizing  on the training points only, and it is also ineffective against adversarial examples.  

\begin{figure}[htb]
    \centering 
    \includegraphics[width=0.24\linewidth]{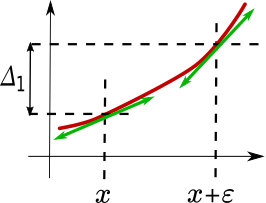}
    \includegraphics[width=0.24\linewidth]{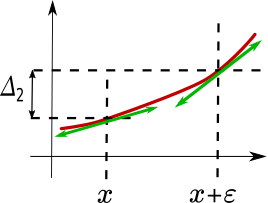}
    \includegraphics[width=0.24\linewidth]{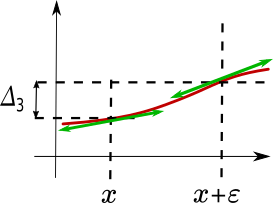}
    \includegraphics[width=0.24\linewidth]{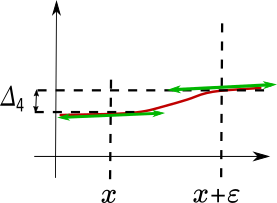}
    \caption{ \small Figure showing the ideal progression of the training for one training point $x$, and associated adversarial points. At first, the difference $\Delta = f(x+\varepsilon) - f(x) $ of ordinates and  gradients norm at $x$ and $x+\varepsilon$ are high. During the course of the training, $\Delta$ decreases, making the value at $x$ and $x+\varepsilon$ closer. The gradients' norm also decrease. In the end, we have almost the same value for $f$ at those points, and the gradients are nearly 0. In those conditions, we have $\Delta_1>\Delta_2>\Delta_3>\Delta_4\approx0$, and the variation between $x$ and $x+\varepsilon$ is then very small.  }
    % Then There exist a point $\z$ which is a convex combination of   $\x$ and $\y$ 
    \label{fig:training_progression}
\vspace{-0.7em}
\end{figure}

Better results are obtained when we penalize the gradients on the training set and also on the adversarial example near the training set. Doing so each training point is associated with an adversarial example, as in classical adversarial training, and we penalize the gradient on both these points. Let $\x$ be a point of the training set and $\x_{adv}$ the adversarial example computed from $\x$ using the  FGSM. Let us assume that we train the classifier for long enough so that $\|\nabla_\x f_i(\x)\|_1 \approx 0$ and $\|\nabla_\x f_i(\x_{adv})\|_1 \approx 0$ for all $i = 1,\dots,c$, since those two points are close enough, the variation of $f$ on the segment joining $\x$ and $\y$ is so small that the class does not change. The ideal training progress is presented in figure \ref{fig:training_progression}. See section \ref{sec:expe2} for comparison of the results of between classical adversarial training, and the new variant introduced in this paper.
The loss function used is:

\vspace{-0.7em}
\begin{equation}\label{eq:loss_gp}
\small
\setstackgap{L}{25pt}
  \alignLongstack{
    \mathcal{L}_{gp}(\x,\y) =& \displaystyle \mathcal{L}(f(\x),\y) + \lambda \sum_{i=1}^c \|\nabla_\x f_i(\x)\|_1 \\
     =&\displaystyle\mathcal{L}(f(\x),\y) + \lambda \sum_{i=1}^c \sum_{j=1}^d |J_{ij}| \\
     =&\displaystyle\mathcal{L}(f(\x),\y) + \lambda \|J\|_{1,1},
    }
\end{equation}

\noindent where $d$ denotes the dimension of the input image, $c$, the number of neurons on the output layers of the neural network \textit{i.e}. the number of classes, $\lambda$, the penalization parameter, $J_{ij} = \frac{\partial f_i(\x)}{\partial x_j } $, the Jacobian matrix of $f$ and $\|.\|_{1,1}$, the entry wise $L_{1}$-norm.

\begin{table}[htb]
\footnotesize
\begin{center}
    \begin{tabular}{|c|c|c|}
    \hline
    model A       & model B       & model C \\ \hline
    conv(64,8,2)  & conv(128,3,1) & FC(512) \\
    conv(128,6,2) & conv(64,3,2)  & FC(256) \\
    conv(128,5,1)       & FC(128)       & FC(128) \\
    FC(10)        & FC(10)        & FC(10)  \\ \hline
      710,218 & 1,460,938 & 567,434\\ \hline
    \end{tabular}
    \vspace{-1em}
\end{center}
\caption{\label{tab:models} \small Description of the models used in this paper: conv($nf,k,s$) represents a convolutional layer with $nf$ filters, of size $k\times k$ applied with a stride $s$. FC($nn$) represents a fully connected layer with $nn$ neurons. All activations are ReLU except the output layer. The numbers in the last line represent the number of parameters.}% \textcolor{red}{ajouter le nombre de paramètres}} 
\end{table}
\vspace{-0.7em}
%---------------------------------------------------------------------------------------------
\section{Experimental Results}
%---------------------------------------------------------------------------------------------
\vspace{-0.7em}
All experiments here are built upon 3 models (referred as A,B and C for simplicity) that are described in table \ref{tab:models}. The first experiments is our proof of concept. It shows that the penalization helps the defense.  Since alone it's still not enough to propose a robust model, the second experiment explores the coupling with adversarial training. 

\subsection{Proof of concept \label{sec:expe1}}
The goal of this experiment is to show that penalizing the gradient improves a lot the robustness of the model while keeping the efficiency on the clean data, and the more we penalize, the more the effect is visible. In our tests, we observe that we double (at least) the number of correctly classified adversarial examples. 

We attack the model using the FGSM in a white-box setup in which the attacker has access to the parameters of the model and to the test set.  The value $\varepsilon = 0.3$ is  chosen as the intensity of the pixel wise perturbation allowed to the attacker as it is often the case for the data used in this experiment. Several values of $\lambda$ are used, allowing to have a notion of the influence of that hyper-parameter on the performance on adversarial samples. The MNIST dataset  and three different architectures are used. The training and testing split of Keras is used, and the training set is then split into two groups, one of 55000 used to train models and the remaining 5000 are used as a validation set. Models are trained until their accuracy on the  validation set stop increasing (difference of accuracy between 2 epochs less than $0.0001$) after 10 epochs, or until 100 epochs of training on the clean data is achieved. This process is repeated 10 times, the mean value and the standard deviation are recorded in table \ref{tab:tab1}.  The program always terminated before 100 hundred epochs were reached, around the 20$^{th}$ epoch.
The optimization method used is Adam with a learning rate of 0.001, $\beta_1=0.9$, $\beta_2=0.999$ and $\epsilon=10^{-8}$, a label smoothing of 0.1 was also used.   %The accuracy on the adversarial validation set is computed every epoch. The adversarial validation set is generated using the validation set and the models' parameters. 
\begin{center}
\begin{table}[htb]
\scriptsize
\begin{tabular}{|l|l|l|l|l|l|l|}
\hline
\multicolumn{2}{|l|}{$\lambda$}       & 50     & 25 & 10 & 1 & 0.1  \\ \hline
\multirow{2}{*}{A} & clean & $99.37 \pm 0.04$ &  $99.36 \pm 0.03$    & $99.38 \pm 0.04$    &  $99.35\pm0.03  $  & $99.35\pm 0.06 $ \\ \cline{2-7} 
                         & adv  &  $45.09\pm5.36 $  & $44.96\pm 4.44$  & $46\pm10.16 $    & $28.83\pm 9.34$    & $22.64\pm 5.02$ \\ \hline
\multirow{2}{*}{B} & clean & $98.99\pm 0.05$  &  $98.99\pm0.05$    &  $98.98\pm 0.07$   &   $98.94\pm 0.07$  & $98.90\pm 0.05$  \\ \cline{2-7} 
                         & adv  & $6.35\pm 3.258$  & $4.69\pm1.91$    & $3.69\pm 0.73$    & $3.00\pm 1.14$ & $3.17 \pm 0.82$ \\ \hline
\multirow{2}{*}{C} & clean &  $98.63\pm0.11$    &   $98.67\pm0.05$ & $98.57\pm0.06$    &  $98.55\pm0.05$   &$98.60\pm 0.08$  \\ \cline{2-7} 
                         & adv  &  $29.92\pm14.24$      & $25.76\pm11.96$   &  $18.24\pm9.97$   & $19.99\pm5.63$     & $13.66\pm 6.64$  \\ \hline
\end{tabular}
\caption{
\label{tab:tab1} \small Accuracy on clean test and adversarial test point of model. We observe that the more we penalize, the more robust the model becomes. However, while the gain is significant, it's clearly not enough to call it a robust model.
%\textcolor{red}{explorer des valeurs  plus grandes pour $\lambda$} 
}
\vspace{-1em}
\end{table}
\end{center}

\vspace{-2em}
\subsection{Coupling with adversarial training \label{sec:expe2}}

\begin{table}[htb] 
\scriptsize
\begin{tabular}{|c|c|p{0.7cm}|p{0.7cm}|p{0.7cm}|p{0.7cm}|p{0.7cm}|p{0.7cm}|p{0.7cm}|p{0.7cm}|}
\hline
\multicolumn{2}{|c|}{\multirow{2}{*}{$\varepsilon$}} & \multicolumn{2}{c|}{0.05} & \multicolumn{2}{c|}{0.1} & \multicolumn{2}{c|}{0.2} & \multicolumn{2}{c|}{0.3} \\ \cline{3-10} 
\multicolumn{2}{|c|}{}                     & clean        & adv        & clean        & adv       & clean        & adv       & clean        & adv       \\ \hline
\multirow{2}{*}{A}  & adv\_train     &   99.36 $\pm0.05$           & 98.57 $\pm0.11$ & 99.34 $\pm0.06$        & 97.50 $\pm0.26$          & 99.27 $\pm0.03$  & 96.39 $\pm0.37$          &     99.30 $\pm 0.06$         &    95.72 $\pm 0.19$       \\ \cline{2-10} 
                          & adv\_train\_gp &  99.38 $\pm 0.028$            &  \bf 98.73 $\pm 0.072$          &   99.35 $\pm 0.035$           & 97.45 $\pm 0.30$   &  99.25 $\pm 0.061 $           &  96.63 $\pm 0.30$         &       99.26 $\pm 0.06$         &    \bf 97.60 $\pm 0.34$       \\ \hline
\multirow{2}{*}{B}  & adv\_train     &      98.94 $ \pm 0.079$    &      98.37 $\pm 0.48$      &       99.07 $\pm 0.03$       &     98.22 $\pm 0.61$      &             98.96 $\pm 0.05$ &         98.55 $\pm 0.26$  &       98.85 $\pm 0.05$       &     98.67 $\pm 0.07$      \\ \cline{2-10} 
                          & adv\_train\_gp &     99.05 $\pm 0.05$     &     98.18 $\pm 0.15$  &        99.06 $\pm 0.05$      &   97,64 $\pm 1.04$    &        98.95 $\pm 0.02$      &     98.30 $\pm 0.49$     &          98.84 $\pm 0.07$      &     97.80 $\pm 1.11$      \\ \hline
\multirow{2}{*}{C}  & adv\_train     &     98.98 $\pm 0.06$         &     96.26 $\pm 0.22$       &      98.68 $\pm 0.06$        &     93.56 $\pm 0.47$      &      98.59 $\pm 0.07$        &    93.24 $\pm 0.48$       &       98.48 $\pm 0.1$       &    95.52 $\pm 0.74$       \\ \cline{2-10} 
                          & adv\_train\_gp &      98.79 $\pm 0.04$        &     96.36 $\pm 0.11$       &       98.87 $\pm 0.07$       &     \bf 94.56 $\pm 0.16$      &       98.69 $\pm 0.09$       &       93.73 $\pm 0.75$    &   98.42 $\pm0.08$            & \bf 96.62 $\pm0.51$          \\ \hline
\end{tabular}
\caption[]{\label{tab:expe2} \small Table showing the performance of models A, B and C, trained with adversarial training and adversarial training plus gradient penalty ($\lambda=10$)\footnotemark.
 We can see that, the {adv\_gp} performs better for model A and $\varepsilon=0.3$, and is sensibly equal for other values of $\varepsilon$. The large standard deviation values might be an indication that the training was still in progress when epoch $100^{th}$ epochs was reached.}
 \vspace{-1em}
\end{table}
\footnotetext{The loss function used here is not exactly the one proposed in \ref{eq:loss_gp}. Instead of $\|J\|_{1,1}$, $\sum_{j=1}^d|\sum_{i=1}^c\frac{\partial f_i(\x)}{\partial x_j}|$ is used due to time constraints.}

The previous experiments show that our penalization has an interesting effect on the robustness of the models. Now we explore the coupling with adversarial training, and the hope is that the penalization has the same impact, in order to obtain more robust models than adversarial training alone. 
We train the same model using adversarial training with the classical loss function and with the gradient penalty in addition to that loss. We train for one hundred epochs. %\textcolor{red}{remind, the concept of adversarial training}.
The adversarial training consists in adding adversarial examples to the original training set to make the models more robust. Typically, the FGSM is used to generate those adversarial examples due to its practicality. So it will be used in the following experiments with a clipping to make sure that adversarial examples remains in $\mathcal{X}$. During training, for each image of a batch, an adversarial image is generated and added to the batch with the same target as the original image associated. A label smoothing of 0.1 is also used for this experiment. 

Table \ref{tab:expe2} shows results for $\lambda=10$. It turns out that performances are not as impressive as they were without adversarial training, even though for some settings, the penalization helps. One reason could be that $\lambda$ is not high enough. However for computing time reason, we could not conduct the same set of experiments with different values at the same level of care. Hence we propose some partial results to argue our point in table where we see that increasing $\lambda$ provides better results. We note also that we increased the number of epoch up to 150.

\begin{table}[ht]
\centering
\scriptsize
    \begin{tabular}{|c|p{0.85cm}|p{0.85cm}|p{0.85cm}|p{0.85cm}|p{0.85cm}|p{0.85cm}|}
        \hline
         $\lambda$ & 0 & 10 & 50 & 100 & 200 & 1000 \\ \hline
         clean & 98.48 $\pm 0.10$ & 98.42 $\pm 0.08$  &  98.49 $\pm 0.06$   &   98.59 $\pm 0.10$  & 98.60 $\pm 0.11$   &    98.62 $\pm 0.07$  \\ \hline
         adv &  95.52 $\pm 0.74$  & 96.62 $\pm 0.51$  & \bf 97.25 $\pm 0.05$  & 96.88 $\pm 0.51$  & 96.71 $\pm 0.19$ & 94.65  $\pm 0.79$\\
         \hline
    \end{tabular}
    \caption{\label{tab:expe3} \small For model C, we fixed $\varepsilon = 0.3$ and different $\lambda$, on the same setting as previous experiment. $\lambda = 0$ refers to the adversarial learning alone. We observe improvement when $\lambda$ increases, until it fails when too high.}
    \vspace{-1em}
\end{table}

\section{Conclusion and future work}
\vspace{-0.7em}
In this paper we propose to improve deep neural networks' robustness to adversarial examples by adding a penalization term.
We show that penalizing the gradients of the output with respect to input, or in other words the Jacobian, can have an effect defending against adversarial but does not suffice. Coupled with adversarial training to give $L_1-$norm double backpropopagation adversarial defense, we provide good hints that it is still a good approach.  
 In future work we will extend the experimental part to more sets of parameters and other datasets.

% ****************************************************************************

% BIBLIOGRAPHY AREA

% ****************************************************************************

\begin{footnotesize}

\bibliographystyle{unsrt}

\bibliography{biblio}

\end{footnotesize}

% ****************************************************************************

% END OF BIBLIOGRAPHY AREA

% ****************************************************************************

\end{document}